\pdfoutput=1
\documentclass[11pt]{article}
\usepackage{amssymb}
\usepackage{graphicx}
\parskip \baselineskip

\newcommand{\N}{\mathbb{N}}

\newcommand{\Z}{\mathbb{Z}}

\newcommand{\ol}{\overline}

\newcommand{\ra}{\rightarrow}
\newcommand{\Ra}{\Rightarrow}
\newcommand{\LRa}{\Leftrightarrow}

\begin{document}

\title{Compression with wildcards:\\
 From CNFs to orthogonal DNFs by imposing \\
the clauses one by one}

%\author{Marcel Wild}
\author{Marcel
Wild\\[3pt]
Department of Mathematical Sciences,
University of Stellenbosch\\
 Private Bag X1, Matieland 7602, South
Africa}

\date{}
%\begin{document}
\maketitle

\begin{quote}
A{\scriptsize BSTRACT}: {\footnotesize We present  a novel technique for converting a Boolean CNF into an orthogonal DNF, aka exclusive sum of products. Our method (which will be pitted against a hardwired command from Mathematica) zooms in on the models of the CNF by imposing its clauses one by one. Clausal Imposition invites parallelization, and wildcards beyond the common don't-care symbol compress the output. The method is most efficient for few but  large clauses. Generalizing clauses one can in fact impose superclauses. By definition, superclauses are obtained from clauses by substituting each positive litereal by an arbitrary conjunction of positive literals.}
\end{quote}

\section{Introduction}

In his 1979 landmark paper [1] Leslie Valiant shifted attention from the SAT problem to the \#SAT problem, i.e. to the task of calculating the exact cardinality of the model set $\mbox{Mod}(\varphi) \subseteq \{0,1\}^w$ of a given Boolean formula $\varphi = \varphi (x_1, \cdots, x_w)$. He showed that many \#SAT problems are so-called \#$P$-hard which roughly speaking means  they are at least as difficult as NP-hard problems. Even problems for which SAT is trivial can be \#P-hard, such as \#DNFSAT.

The ALLSAT problem for $\varphi$, our article's topic, extends \#SAT in so far as not just the number $|\mbox{Mod}(\varphi)|$ is required but the models themselves.  Since $|\mbox{Mod}(\varphi)|$ can be exponential in the input length, one commonly\footnote{A nice account of the various kinds of 'polynomial' enumerability can be found in [2].} regards the ALLSAT problem as {\it solvable} when the enumeration of Mod$(\varphi)$ can be achieved in {\it polynomial total time}. It turns out that some classes ${\cal C}$ of Boolean formulas whose \#SAT problem is \#P-hard nevertheless have a solvable ALLSAT-problem, e.g. the class of all Boolean DNF's.

Unfortunately,  a one-by-one enumeration of $\mbox{Mod}(\varphi)$ (even if polynomial total time) may take centuries when $|\mbox{Mod}(\varphi)|$ gets large. To get a  glimpse of how to mend the problem let
 $\varphi_0: \ \{0,1\}^8 \ra \{0,1\}$ be defined by $\varphi_0 (x_1, \cdots, x_8) = x_3 \vee \ol{x}_4\vee x_6$. Then enumerating the model set Mod$(\varphi_0)$ one-by-one forces us to list 224 length 8 bitstrings. It is more economic to have a representation like

(1) \quad $\mbox{Mod}(\varphi_0) = (2, 2,{\bf 1}, {\bf 2}, 2, 2, 2, 2) \uplus
 (2,2, {\bf 0}, {\bf 0},2, 2, 2, 2)\uplus  (2,2, {\bf 0}, {\bf 1},2, {\bf 1}, 2, 2) $
 
 Although this expression may be self-explanatory for many readers, we will formalize it  in Subsection 1.1. (Always $\uplus$ means {\it disjoint} union.) In 1.2 follows a very brief history of exclusive sums of products, and 1.3 will be the detailed Section break-up. 
 
 {\bf 1.1} A $012$-{\it row} $r$ of length $w$ is what is often called a {\it partial variable assignment}, i.e. a map $r:S\to \{0,1\}$, where $S$ is a subset of the variable set $\{x_1,\ldots,x_w\}$. For instance, take $r:\{x_3,x_4,x_6\}\to\{0,1\}$ defined by $r(x_3)=0,\ r(x_4)=r(x_6)=1$. However, we prefer to {\it list the indices} of the variables which are mapped to 0 and 1 respectively. Thus $zeros(r):=\{3\}$ and $ones(r):=\{4,6\}$. Although this uniquely defines the map $r$ we additionally put $twos(r):=\{1,2,5,7,8\}$. The most visual way (which we adopt throughout) is to simply write $r=(2,2,0,1,2,1,2,2)$ and call this vector a 012-{\it row}. Thus '2' is just an alternative notation for the common don't-care symbol '$*$'. Rather than thinking of a map (=partial variable assignment) we will identify $r$ with the set of bitstrings obtained by freely substituting 0 or 1 for each symbol 2. Thus $|r|=32$. The {\it associated term} for $r$ is $T(r):=\ol{x_3}\wedge x_4\wedge x_6$. Vice versa $r(T)$ is the 012-row coupled to the term $T$. These correspondences are mutually inverse. 
 
 %Strictly speaking a $012$-row $r$ with twos$(r) = \emptyset$ is not quite a bitstring; it is the singleton whose only element is the bitstring.
 
 An {\it exclusive sum of products (ESOP)} for a Boolean function $\varphi$ is a representation of $Mod(\varphi)$ as a disjoint union of 012-rows. Any ESOP of $\varphi$ immediatly yields a DNF $\psi$ which is equivalent to $\varphi$ and which is {\it orthogonal} in the sense that any conjunction of distinct terms is unsatisfiable. In fact $\psi$ is just the disjunction of all terms $T(r)$ where $r$ ranges over the 012-rows constituting the ESOP of $\varphi$. Conversely each orthogonal DNF (ODNF) immediately yields an ESOP. In the literature ESOP is often used synonymous with ODNF, but we stick to the meaning ESOP = disjoint union of 012-rows.

{\bf 1.2} In the 80's ESOPs were often used for network reliability analysis and the starter $\varphi$ was usually in DNF format. Abraham's way [3] to make the terms of $\varphi$ one by one orthogonal to the 'ODNF  obtained so far' was influential. His pattern, which we later on shall visualize as 'Abraham's Flag', features in Theorem 7.1 of [4] and also in [5, Section 3.2]. Both references are recommended for a deeper look at ODNFs in the 80's.
The 90's were dominated by binary decision diagrams (=BDDs). The numerical experiments in [5] leave no doubt that BDDs were superior to the methods of the 80s. The new millenium saw the rise of Davis-Putnam-Logeman-Loveland (DPLL) which continues to be the leading framework (based on the so called resolution method) to settle the satisfiability of a CNF. It later turned out [5] that DPLL can also be applied to obtaining ESOPs even faster than with BDDs.

An ESOP likely is the single most convenient representation of a Boolean function $\varphi$. ESOPs have been used for many purposes, e.g. recently for quantum computing [6]. Among the various ways to obtain an ESOP of $\varphi$ we are mainly interested in the transition $CNF\to ESOP$ and shall discuss three  methods. The first is based on BDDs and the second is Pivotal Decomposition (based on variable-wise branching). Their pros and cons are recalled in Section 2. The novel third method, 'Clausal Imposition', is the core of the present article.
 
{\bf 1.3} Here comes the further Section break-up, starting with Section 3. It discusses two natural ways a
 $012$-row $r$ can relate to 
 $\mbox{Mod}(\varphi)$: We call $r$ {\it feasible} if $r \cap \mbox{Mod}(\varphi) \neq \emptyset$, and {\it final} if $r \subseteq \mbox{Mod}(\varphi)$. These concepts are crucial in the example of  Section 4. Here $\varphi$ is a CNF with four clauses and we gradually zoom in to $\mbox{Mod}(\varphi)$ by imposing the clauses one after the other. This  clausal
 $n$-algorithm (alternatively: Clausal Imposition\footnote{In previous publications related methods were named 'principle of exclusion' or 'clause-wise branching'. With hindsight, 'Clausal Imposition' fits best.}) is in stark contrast to Pivotal Decomposition. It actually embraces Abraham's Flag but now applied to clauses rather than terms. There are other differences as well, such as the $n$-wildcard which goes beyond the classic don't-care symbol '2'. 
 
  So far we spoke about SAT, \#SAT and ALLSAT (mainly ESOP). But there also is OPTIMIZATION. In Section 5 we tentatively float the idea to use Clausal Imposition for finding {\it all} optimal solutions to a 0-1-program (once {\it one} is found), or to tackle {\it multi}objective 0-1-programs. Admittedly, Section 5 is more speculative and less streamlined than the other Sections.
  
Section 6 is devoted to numerical experiments. In the first set of experiments we pit the clausal $n$-algorithm against some hardwired Mathematica competitor (called  {\tt BooleanConvert}), i.e. we evaluate their running times and the achieved compression when transforming random CNFs to ESOPs. The outcome, in a nutshell, is as follows. Clausal Imposition excels for few but long clauses, yet gives way to {\tt BooleanConvert} for many but short clauses. That is unless the many clauses are
implicit, i.e. encrypted in superclauses (as evidenced by our second set of experiments). A third set of experiments combines the {\tt LinearProgramming} command of Mathematica with Clausal Imposition in the spirit of Section 5.
Section 7 recalls past, and forecasts future variations of the clausal $n$-algorithm. Specifically we look at Horn CNF's (and its many subcases) and at 2-CNFs.
  
While Sections 1 to 7 are comparatively 'easy reading', Sections 8 to 10 have a more technical slant and formal Theorems and Corollaries only start to surface here. Specifically, in order to find a common hat for both Clausal Imposition and Pivotal Decomposition (and for potential future schemes), Section 8 introduces the notion of a 'row-splitting mechanism' and proves a Master Theorem. In Section 9 we view Pivotal Decomposition as a row-splitting algorithm and establish three  Corollaries of the Master Theorem.  For instance, it follows at once from Corollary 1 that any algorithm for $DNF\to ESOP$ yields an algorithm for $CNF\to ESOP$.
 Section 10 views Clausal Imposition as a row-splitting mechanism and invokes the Master Theorem to give a theoretic assessment of the clausal $n$-algorithm  informally introduced  in Section 4.

\section{\bf ESOPs from BDDs, respectively Pivotal Decomposition}

We assume a basic familiarity with binary decision diagrams (BDD's), as e.g. provided by Knuth [7, Sec. 7.1.4]. Section 2 prepares the reoccurence of BDD's and Pivotal Decomposition later on.

{\bf 2.1} Consider the Boolean function $\varphi_1 : \{0,1\}^5 \ra \{0,1\}$ that is defined by the BDD in Figure 1.
Whether a bitstring $u$ belongs to Mod$(\varphi_1)$ can be decided by scanning $u$ as follows. The dashed and solid lines descending from a node labelled $x_i$ are chosen according to whether the $i$-th component $u_i$ of $u$ is $0$ or 1. Thus, in order to decide whether $u = (u_1, u_2, u_3, u_4, u_5) = (0,1,0,1,0)$ belongs to Mod$(\varphi_1)$ we follow the dashed line from the root ($=$ top node) $x_1$ to the node $x_2$ (since $u_1 =0$). Then from $x_2$ with the solid line to $x_4$ (since $u_2 =1$), then from $x_4$ with the solid line to $x_5$ (since $u_4 = 1$), then from $x_5$ with the dashed line to the leaf $\perp$ (since $u_5 =0$). Because $0$ means 'False' (and $1$ means 'True') we conclude that $u\not\in Mod(\varphi_1)$.

For each Boolean function $\varphi$, and each given variable-ordering, there is exactly one BDD of $\varphi$; see [7, p.204]. Thus modulo a fixed variable-ordering  the notation $BDD(\varphi)$ is well-defined.

\begin{center}
\includegraphics[scale=0.5]{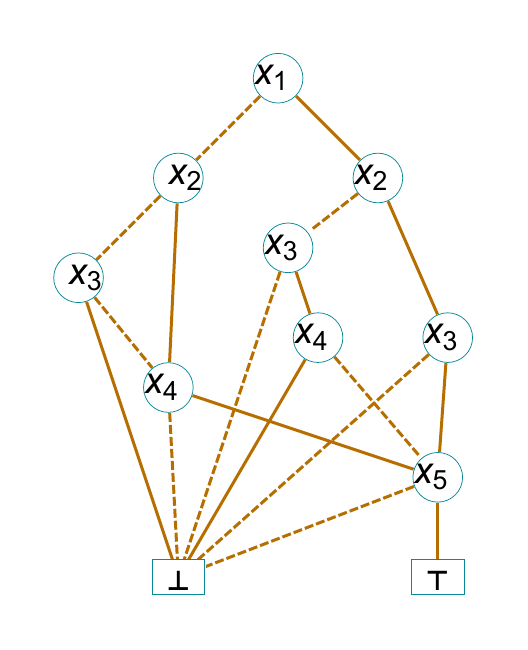} 

{\sl Figure 1: This BDD defines a Boolean function $\varphi_1$ from $\{0,1\}^5$ to $\{0,1\}$.}
\end{center}

{\bf 2.2} Notice that the value $u_3=0$ in the bitstring $u=(0,1,0,1,0)\in Mod(\varphi_1)$ is irrelevant since there is an {\it index-gap} between $x_2$ and $x_4$ in the accepting path of $u$. Therefore $(0,1,2,1,0)\subseteq Mod(\varphi_1)$. As is well known, in any $BDD(\varphi)$ the paths from the root to $\top$ match the $R$ products (=012-rows) of a {\it BDD-induced} ESOP of $\varphi$. The more and the larger the index-gaps in $BDD(\varphi)$, the more $2$'s occur in the 012-rows, and hence the higher is the compression of $Mod(\varphi)$ provided by the ESOP. Evidently each BDD-induced ESOP of $\varphi$ can be generated in time linear in $R$ and the size of $BDD(\varphi)$. Calculating the mere cardinality $|Mod(\varphi)|$ can be done in time linear in the size of the BDD (irrespective of $R$). In plain language, it works in the blink of an eye even when the BDD is very large.

{\bf 2.3} Recall that the {\it Hamming-weight} of a bitstring is the number of 1-bits it contains. Here comes another benefit of BDD's. 
If $\varphi$ is given by a BDD, what is the best way of counting respectively enumerating all models of fixed Hamming-weight $k$? As to counting,  Knuth  [7,p.206] shows the way. As to enumeration, let us introduce the wildcard $(g_t,g_t,\ldots,g_t)$ which means 'exactly $t$ many 1's here'. For instance $(g_2,g_2,g_2,0,1,g_3,g_3,g_3,g_3,g_3)$ by definition comprises the ${{3}\choose{2}}{{5}\choose{3}}=30$ length 10 bitstrings having exactly two 1's in the part $(g_2,g_2,g_2)$ and exactly three 1's in the part  $(g_3,g_3,g_3,g_3,g_3)$. These bitstrings hence have Hamming-weight 2+1+3=6. According to [8], if $BDD(\varphi)$ is given, then the $\varphi$-models of any fixed Hamming-weight  can be enumerated in a compressed way, and in time polynomial in $w$ and $|BDD(\varphi)|$.

{\bf 2.4} As glimpsed above and by  other reasons [7, Sec. 7.1.4],  BDD's are great once you have them. Unfortunately calculating BDD's may trigger intermediate tree-like structures much larger than the final BDD. There is also the (not to be underestimated) pedagogical point of view: Although Knuth [7,p.218] does a better job than many in explaining how $BDD(\psi_1\wedge\psi_2)$ (or $BDD(\psi_1\vee\psi_2)$, etc) arises from $BDD(\psi_1)$ and  $BDD(\psi_2)$, the following two methods are arguably easier to grasp, let alone  to program. One is Clausal Imposition,
the main topic of our article (starting in Section 4).

{\bf 2.5} The other one is {\it Pivotal Decomposition}. Before we trace its origin, we illustrate it on a toy example. So consider the Boolean function

(2)\quad $\varphi_2(x_1,x_2,x_3,x_4):=[ (x_2\vee x_3)\to x_1]\ \wedge\ [(x_3\wedge x_4)\to(x_1\wedge x_2)]$

 For starters, $Mod(\varphi_2)$ is a subset of $\{0,1\}^4=(2,2,2,2)=:r_0$ (see Table 1). If by whatever reason we decide to branch on the pivot $x_1$ (see $r_1,r_2$), then $Mod(\varphi_2\wedge x_1)=Mod((x_3\wedge x_4)\to x_2)\subseteq r_1$ (set $x_1=1$ on the right hand side of (2)), and $Mod(\varphi_2\wedge\ol{x_1})=Mod((\ol{x_2}\wedge \ol{x_3})\wedge (\ol{x_3}\vee \ol{x_4}))\subseteq r_2$ (set $x_1=0$).
 Clearly $r_1\cap r_2=\emptyset$, which by induction will establish that the final 012-rows will be disjoint, i.e. constitute an ESOP.
 
 \begin{tabular}{l|c|c|c|c|c}
 	& $x_1$ & $x_2$ & $x_3$ & $x_4$  & \\ \hline
 	& & & & &  \\ \hline
 	$r_0=$ &  2 & 2 & 2 & 2 & next branch on $x_1$ \\ \hline
 	& & & & &  \\ \hline 
	$r_1=$ &  {\bf 1} & 2 & 2 & 2 & $(x_3\wedge x_4)\to x_2$, next branch on $x_4$ \\ \hline
 	$r_2=$ &   {\bf 0} & 2 & 2 & 2 & 
 	$(\ol{x_2}\wedge \ol{x_3})\wedge (\ol{x_3}\vee \ol{x_4})$, next branch on $x_3$ \\ \hline
 		& & & & &  \\ \hline 
 	$r_3=$ &  1 & 2 & 2 &  {\bf 1} & $x_3\to x_2$, next branch on $x_3$ \\ \hline	
 	$r_4=$ &  1 & 2 & 2 &  {\bf 0} & $0\to x_2\ \equiv\ 1$, save $r_4$ \\ \hline
 		$r_2=$ & 0 & 2 & 2 & 2 & 
 	$(\ol{x_2}\wedge \ol{x_3})\wedge (\ol{x_3}\vee \ol{x_4})$, next branch on $x_3$ \\ \hline
 			& & & & &  \\ \hline
 		$r_5=$ &  1 & 2 &  {\bf 1} & 1 & $1\to x_2\ \equiv\ x_2$, next put $x_2=1$ \\ \hline
 			$r_6=$ &  1 & 2 &  {\bf 0} &1 & $0\to x_2\ \equiv\ 1$, save $r_6$ \\ \hline
 	$r_2=$ &  0 & 2 & 2 & 2 & $(\ol{x_2}\wedge \ol{x_3})\wedge (\ol{x_3}\vee \ol{x_4})$, next branch on $x_3$ \\ \hline
 	
 					& & & & &  \\ \hline
 					
	$r_7=$ &  1 & 1 &  1 &1 & save $r_7$ \\ \hline
	$r_2=$ &  0 & 2 & 2 & 2 & 
 		$(\ol{x_2}\wedge \ol{x_3})\wedge (\ol{x_3}\vee \ol{x_4})$, next branch on $x_3$ \\ \hline
 		
 			& & & & &  \\ \hline

 		$r_8=$ &  0 & 2 &  {\bf 1} & 2 & 
 		       $(\ol{x_2}\wedge 0)\wedge (0\vee \ol{x_4})\ \equiv\ 0$, kill $r_8$    \\ \hline
 					$r_9=$ &  0 & 2 &  {\bf 0} & 2 & 			
 	            $(\ol{x_2}\wedge 1)\wedge (1\vee \ol{x_4})\ \equiv\ \ol{x_2}$, next put $x_2=0$ \\ \hline
 	            
 	            	& & & & &  \\ \hline
 	            	
 	            		$r_{10}=$ &  0 & 0 &   0 & 2 & save $r_{10}$ \\ \hline

 \end{tabular}
 
 Table 1: {\sl Snapshots of the working stack during Pivotal Decomposition.}
 
Always focusing on the top row of the LIFO (=Last In First Out) {\it working stack} one can decide on the fly which pivot to branch on.
To the right of each 012-row $r_i$ we wrote the formula which still needs to be 'imposed' on $r_i$. This formula might be equivalent to $1$ (thus a tautology), in which case $r_i$ is {\it final}; it  gets removed from the working stack and is saved somewhere else. Or the formula might be equivalent to $0$ (i.e. unsatisfiable), in which case $r_i$ gets killed without a trace.
 Altogether there are four final rows which hence yield the ESOP $Mod(\varphi_2)=r_4\uplus r_6\uplus r_7\uplus r_{10}$. Using other pivots for branching (e.g. first $x_2$, then $x_1$ if $x_2=1$, respectively $x_3$ if $x_2=0$), we invite the reader to check that
$Mod(\varphi_2)$ gets rendered as $(1,1,2,2)\uplus (1,0,1,0)\uplus (2,0,0,2)$.
 
{\bf 2.5.1} The terminology 'Pivotal Decomposition' (adapted from [5]) is better than 'variable-wise branching' (the latter being used already in many other contexts). Pivotal Decomposition can be used to enumerate $Mod(\varphi)$ for any Boolean formula $\varphi$. Indeed, as illustrated in the toy example upon branching on any pivot, by induction at least one of the branches delivers a satisfiable formula. In general a  SAT-solver is used to decide which branches are good. We see that Pivotal Decomposition generally leads to a compressed enumeration of $Mod(\varphi)$, i.e. to a {\it proper} ESOP as opposed to one-by-one enumeration. If the SAT-solver has polynomial complexity\footnote{Loosely speaking, this takes place iff $\varphi$ is on the good side of the Schaefer dichotomy (google). Specifically, the CNF's $\varphi$
	for which 'polynomial total time' can be improved to 'polynomial delay' are classified in [9]. For instance, Horn functions are of the latter kind. The author dares to argue that in practise compression matters more than the label 'polynomial delay'. Moreover compressed enumeration does not discriminate against $\varphi$'s on the wrong side of the Schaefer dichotomy.} then the enumeration of $Mod(\varphi)$ runs in polynomial total time. 
 
 {\bf 2.5.2} How does Pivotal Decomposition compare to BDD's? One major advantage of Pivotal Decomposition is that the choice of pivots can be decided on the fly, whereas for BDD's the variable order has to be fixed at the beginning. This {\it PivDec-induced} ESOP is therefore often shorter (see [5] and also Section 6) than the BDD-induced ESOP defined in 2.2.

   {\bf 2.5.2.1} In particular, if $\varphi$  is a DNF then Pivotal Decomposition can dispense with SAT-solving altogether  because all occuring subformulas are again DNF's (possibly tautologies or unsatisfiable), i.e. DNFs are 'hereditary'. Furthermore,  for DNF's the pivots can be chosen not just on the fly, but according to efficient heuristics [5]. Corollary 2 in Section 9 lifts Pivotal Decomposition and DNF's to a more abstract framework, but one which still relies on 'heritance'.
   
   {\bf 2.5.2.2} If $\varphi$ is in CNF format (i.e. a conjunction of clauses $C_1,C_2,\ldots$), then one cannot dispense with a SAT-solver, yet there still is a pivot selection strategy that springs to mind (and possibly was observed before). Namely, if $r$ is the top row of the working stack and $i\in twos(r)$, then let $f_1(i)$ be the number of clauses $C_j$ with literal $x_i$ such that $C_j$ is not yet satisfied\footnote{By definition we say that $C_j$ is {\it satisfied by r} if $C_j$ either contains a literal $x_k$ with $k\in ones(r)$, or a literal $\ol{x_k}$ with $k\in zeros(r)$.} by $r$. Similarly 
let $f_0(i)$ be the number of clauses $C_j$ with literal $\ol{x_i}$ such that $C_j$ is not yet satisfied
by $r$. It seems a good idea to pick the pivot $i$ that maximizes $f_1(i)+f_0(i)$.

\section{Feasibility and finality of $012$- and $012n$-rows}

Let $\varphi : \{0,1\}^w \ra \{0,1\}$ be any Boolean function. 
An $012$-row $r$ of length $w$ is {\it $\varphi$-feasible} (or just {\it feasible}) if $r \cap \, Mod(\varphi) \neq \emptyset$. Hence the feasibility of $r$ amounts to say that  $T(r) \wedge \varphi$ is satisfiable.
A {\it (proper) feasibility test} is an algorithm which, when fed with a $012$-row $r$, produces an answer {\it yes} or {\it no}, namely '${\it yes}
\LRa\ r\ \hbox{\it is feasible}$' (thus '${\it no}\ \LRa\ r\ \hbox{\it is infeasible}$)'. Unless $\varphi$ has special shape one needs a general-purpose SAT-solver to decide the feasibilty of $r$.
We speak of a {\it weak feasibility test} if it only holds that '${\it yes}\Ra\ r\ \hbox{\it is feasible}$'. 

We say that $r$ is {\it final}  if $r\subseteq Mod(\varphi)$.
A  {\it finality test}  is an algorithm which, when fed with a $012$-row $r$, produces an answer {\it yes} or {\it no} akin to the above: ${\it yes}\LRa\ r\ \mbox{\it is final}$. Weak finality tests will not be relevant in our article. (Weak finality tests are not relevant in our article.)
Of course for all 012-rows it holds that  'final $\Ra$ feasible', and for 01-rows it holds that 
'final $\LRa$ feasible'.

{\bf 3.1} The difficulty to decide the feasibility or finality of $r$  depends  on how $\varphi$ is presented. In this article
 $\varphi$ is mainly given as a CNF. To fix ideas, consider this Boolean function $\varphi_3:\{0,1\}^8\to\{0,1\}$:

(3) \quad $(\ol{x}_1\vee\ol{x}_2\vee \ol{x}_3\vee x_4)\wedge (x_4\vee x_5\vee x_6\vee\ol{x}_7\vee\ol{x}_8)
\wedge  (x_2\vee x_3\vee x_4\vee x_5\vee\ol{x}_6\vee\ol{x}_7)
\wedge (\ol{x}_5\vee \ol{x}_8)$

A 012-row $r\subseteq\{0,1\}^8$ that wants to be $\varphi_3$-feasible must be feasible for each of the four clauses of $\varphi_3$ individually. However, this is not sufficient. For instance, $r=(2,2,2,0,2,0,1,1)$ is feasible for all of them individually (obvious) but infeasible for $\varphi_3$ (less obvious but witnessed by Table 3). Thus the straightforward  feasibility test for clauses only yields a weak feasibility test for CNF's.

{\bf 3.2} The $n$-wildcard $(n,n,\ldots,n)$ mentioned in 1.3 means 'at least one 0 here'. To spell it out, if there are $t$ symbols $n$, then $2^t-1$ bitstrings are allowed, i.e. all except $(1,1,\ldots,1)$. We will distinguish distinct $n$-wildcards by subscripts. The so obtained {\it 012n-rows} [10] are handy generalizations of 012-rows.
Thus the $012n$-row $r=(2,n_1,n_2,0,1,n_2,0,n_1,n_2)$ by definition is the set of  bitstrings $(x_1,\ldots,x_9)$ which have $x_4=x_7=0$ and $x_5=1$ and $0\in\{x_2, x_8\}$ and $0\in\{x_3, x_6, x_9\}$. The cardinality of $012n$-rows, i.e. the number of bitstrings they contain, is easily calculated, say

 (4)\quad $r=(2,n_1,n_2,0,1,n_2,0,n_1,n_2)\ \Ra \ |r|=2\cdot(2^2-1)\cdot(2^3-1)=42$. 

The definition of $\varphi$-feasibility or $\varphi$-finality of a 012n-row $r$ extends in the expected way, i.e. $r\cap Mod(\varphi)\not=\emptyset$ respectively $r\subseteq Mod(\varphi)$. Notice that say $(n,n,0,1,2)$ is final with respect to  $\ol{x}_1\vee \ol{x}_2\vee x_3\vee\ol{x}_4\vee x_5$ since a whole $n$-wildcard is subsumed by negative literals.

{\bf 3.2.1} The intersection of two 012-rows is either empty or again a 012-row. In fact, the intersection of a 012-row $r$ with a $012n$-row $\rho$ is either empty or otherwise again a $012n$-row. 
For instance:

(5)\quad  $(0,0,0,1,1,2,2,2)\cap (n_1,0,2,n_1,n_2,n_2,n_2,n_1)=(0,0,0,1,1,n_2,n_2,2)$

An empty intersection occurs iff
either a 0 of one row occupies the same position as a 1 of the other, or the position-set of a $n$-wildcard of $\rho$ is wholly contained in $ones(r)$. (The intersection of two proper $012n$-rows is harder to calculate.)

\section{CNF to ESOP by Clausal Imposition: An example}

We turn to (3) and set $\varphi=\varphi_3=C_1\wedge C_2\wedge C_3\wedge C_4$ (the four clauses). Putting 
$Mod_i:=Mod(C_1\wedge..\wedge C_i)\subseteq \{0,1\}^8$
we start with $Mod_0=\{0,1\}^8$ (encoded as $r_0$ in Table 2) and attempt to sieve $Mod_{i+1}$ from $Mod_i$ until we arrive at $Mod_4=Mod(\varphi_2)$. We say that a $012n$-row (briefly: row) $r$ {\it satisfies the clause} $C_j$ if $r\subseteq Mod(C_j)$), i.e. if $r$ is $C_j$-final. As to sieving $Mod_1$ from $Mod_0$, each $u\in Mod_1$ either satisfies $\ol{x}_1\vee \ol{x}_2\vee \ol{x}_3$ or not. If yes, $u\in r_1$ (as defined in Table 2). If no, we must have $x_4=1$, and so $u\in r_2$. Hence $Mod_1=r_1\uplus r_2$. Since by construction $r_1$ satisfies $C_1$ the {\it pending clause} to be imposed upon $r_1$ is $C_2$, in brief $pc=2$. Likewise $r_2$ satisfies $C_1$ but it also happens to satisfy $C_2$ and $C_3$ (because each $(x_1,\ldots,x_9)\in r_2$ has $x_4=1$). Hence $r_2$ has $pc=4$.

 The rows $r_1$ and $r_2$ currently constitute our LIFO {\it working stack}. We hence always focus on its topmost row, akin to Table 1. Thus the next task is to impose $C_2$ upon $r_1$. The {\it clausal}\footnote{This is not standard terminology but it relates to the standard notion of 'implication' in Subsection 7.1.} {\it implication}
 $(x_7\wedge x_8)\to (x_4\vee x_5\vee x_6)$ is equivalent to $C_2=(x_4\vee x_5\vee x_6\vee\ol{x_7}\vee\ol{x_8})$ in (3), but is more handy. A bitstring $u\in r_1$ satisfies that formula if either the {\it premise}  $x_7\wedge x_8$ is violated, or both the premise and {\it conclusion} $x_4\vee x_5\vee x_6$ are satisfied. The former $u$'s are collected in $r_3$, the latter $u$'s in $r_4\uplus r_5\uplus r_6$. As in previous publications we call $r_3$ to $r_6$ the {\it candidate sons} of $r_1$. A  pattern like the boldface $3\times 3$ square within $r_4,\ r_5,\ r_6$ will be called an {\it Abraham 1-Flag}\footnote{In previous articles other names were used, but in view of the comments in 1.2 a reference to Abraham seems most fitting.}. (The 1 in 1-Flag refers to the 1's in the diagonal.) Its benefit is that $(1,2,2)\uplus (0,1,2)\uplus (0,0,1)$, as opposed to $(1,2,2)\cup (2,1,2)\cup (2,2,1)$ is {\it disjoint} without being longer. 

\begin{tabular}{l|c|c|c|c|c|c|c|c|c}
& $x_1$ & $x_2$ & $x_3$ & $x_4$ & $x_5$ & $x_6$ & $x_7$ & $x_8$ & \\ \hline
& & & & & & & &  & \\ \hline
$r_0=$ & 2 & 2 & 2 & 2 & 2 & 2 & 2 & 2 & $pc=1$\\ \hline
& & & & & & & & & \\ \hline
$r_1=$ & $n$ &  $n$ &   $n$ & 2 & 2 & 2 & 2 & 2 & $pc=2$ \\ \hline 
$r_2=$ & 1 & 1 & 1 & 1  & 2 & 2 & 2 & 2 & $pc=4$ \\ \hline
& & & & & & & & &  \\ \hline
$r_3=$ & $n_1$ & $n_1$ & $n_1$ &  2 & 2 & 2 & $n_2$ & $n_2$ & $pc=3$ \\ \hline
$r_4=$ & $n$ & $n$ & $n$ & ${\bf 1}$ & ${\bf 2}$ & ${\bf 2}$ & 1 & 1 & $pc=4$ \\ \hline
$r_5=$ & $n$ & $n$ & $n$ & ${\bf 0}$ & ${\bf 1}$ & ${\bf 2}$ & 1 & 1 & {\it infeasible} \\ \hline
$r_6=$ & $n$ & $n$ & $n$ & ${\bf 0}$ & ${\bf 0}$ & ${\bf 1}$ & 1 & 1 & $pc=3$ \\ \hline
$r_2=$ & 1 & 1 & 1 & 1  & 2 & 2 & 2 & 2 & $pc=4$ \\ \hline
& & & & & & & & &  \\ \hline
$r_7=$ &  $n_1$ & $n_1$ & $n_1$ & 2 & 2 & ${\bf 0}$ & ${\bf n_2}$ & $n_2$ & $pc=4$ \\ \hline
$r_8=$ &  $n$ & $n$ & $n$ & 2 & 2 & ${\bf 1}$ & ${\bf 0}$ & 2 & $pc=4$ \\ \hline
$r_9=$ & $n$ & ${\bf 1}$ & ${\bf n}$ & ${\bf 2}$ & ${\bf 2}$ & 1 &  1 & 0 & {\it final} \\ \hline 
$r_{10}=$ & 2 & ${\bf 0}$ & ${\bf 1}$ & ${\bf 2}$ & ${\bf 2}$ & 1 &  1 & 0 &  {\it final}\\ \hline 
$r_{11}=$ & 2 & ${\bf 0}$ & ${\bf 0}$ & ${\bf 1}$ & ${\bf 2}$ & 1 &  1 & 0 &  {\it final}\\ \hline 
$r_{12}=$ & 2 & ${\bf 0}$ & ${\bf 0}$ & ${\bf 0}$ & ${\bf 1}$ & 1 &  1 & 0 &  {\it final}\\ \hline
$r_4=$ & $n$ & $n$ & $n$ & 1 & 2 & 2 & 1 & 1 & $pc=4$ \\ \hline
$r_6=$ & $n$ & $n$ & $n$ & 0 & 0 & 1 & 1 & 1 & $pc=3$ \\ \hline
$r_2=$ & 1 & 1 & 1 & 1  & 2 & 2 & 2 & 2 & $pc=4$ \\ \hline
& & & & & & & & &  \\ \hline 
$r_4=$ & $n$ & $n$ & $n$ & 1 & 2 & 2 & 1 & 1 & $pc=4$ \\ \hline
$r_6=$ & $n$ & $n$ & $n$ & 0 & 0 & 1 & 1 & 1 & $pc=3$ \\ \hline
$r_2=$ & 1 & 1 & 1 & 1  & 2 & 2 & 2 & 2 & $pc=4$ \\ \hline
& & & & & & & & & \\ \hline 
$r_{13}=$ & $n$ & ${\bf 1}$ & ${\bf n}$ & 0 & 0 & 1 & 1 & 1 & {\it final} \\ \hline
$r_{14}=$ & 2 & ${\bf 0}$ & ${\bf 1}$ & 0 & 0 & 1 & 1 & 1 & {\it final} \\ \hline
$r_2=$ & 1 & 1 & 1 & 1  & 2 & 2 & 2 & 2 & $pc=4$ \\ \hline
\end{tabular}

Table 2: {\sl Snapshots of the working stack of the clausal $n$-algorithm.}

Notice that the candidate son $r_5$ needs to be cancelled since it is infeasible for $C_4=\ol{x}_5\vee \ol{x}_8$ (no $u\in r_5$ satisfies $C_4$). Afterwards the working stack (from top to bottom) consists of $r_3,\ r_4,\ r_6,\ r_2$. We continue by imposing $C_3$, written as $(x_6\wedge x_7)\to (x_2\vee x_3\vee x_4\vee x_5)$, upon $r_3$. The bitstrings $u\in r_3$ violating the premise are collected in $r_7\uplus r_8$. Here the boldface area is a {\it Abraham 0-Flag}. (As to $r_8$, turning the first component of $(n_2,n_2)$ to 0 'frees' the second component, i.e. yields $(0,2)$.) The $u\in r_3$ that satisfy both premise and conclusion are collected in $r_9\uplus r_{10}\uplus r_{11}\uplus r_{12}$. All four rows happen to satisfy $C_4$, whence are contained in $Mod(\varphi_2)$, whence are final in the sense of Section 3. We transfer them from the working stack onto the {\it final stack} in Table 3 (read bottom-up). Now the working stack consists of $r_7,\ r_8,\ r_4,\ r_6,\ r_2$. Imposing the last clause $C_4$  
upon $r_7$ yields $(n_1,n_1,n_1,2,{\bf 0},0,n_2,n_2)\uplus(n,n,n,2,{\bf 1},0,2,{\bf 0})$. These two rows are put on the final stack. Imposing $C_4$ upon $r_8$ yields $(n_1,n_1,n_1,2,n_2,1,0,n_2)$ which is also put on the final stack.

Afterwards the working stack is $r_4,\ r_6,\ r_2$ (Table 2). Imposing $C_4$ upon $r_4$ yields $(n,n,n,1,{\bf 0},2,1,1)$ which is put on the final stack. Imposing $C_3$ upon $r_6$ yields $r_{13}\uplus r_{14}$. Both rows happen to be final. Imposing $C_4$ upon the last row $r_2$ in the working stack yields $(1,1,1,1,n,2,2,n)$ which completes the final stack. Because $Mod(\varphi_3)$ is the disjoint union of the final rows, evaluating them as in (4) yields

(6)\quad $|Mod(\varphi_3)|\ =\ 12+2+3+\cdots + 8+12\ =\ 169$.

\begin{tabular}{l|c|c|c|c|c|c|c|c|c}
 & 1 & 1 & 1 & 1 & $n$ & 2 & 2 & $n$ &$\ 12$ \\ \hline
 & 2 &  0 &   1 & 0 & 0 & 1 & 1 & 1 &$\ 2$  \\ \hline 
 & $n$ & 1 & $n$ & 0  & 0 & 1 & 1 & 1 &$\ 3$  \\ \hline
 & $n$ & $n$ & $n$ &  1 & 0 & 2 & 1 & 1 &$\ 14$  \\ \hline
 & $n_1$ & $n_1$ & $n_1$ & 2 & $n_2$ & 1 & 0 & $n_2$ & $\ 42$ \\ \hline
& $n$ & $n$ & $n$ & 2 & 1 & 0 & 2 & 0 & $\ 28$ \\ \hline
 & $n_1$ & $n_1$ & $n_1$ & 2 & 0 & 0 & $n_2$ & $n_2$ & $\ 42$ \\ \hline
 & 2 & 0 & 0 & 0 & 1 & 1 & 1 & 0 & $\ 2$ \\ \hline
 & 2 & 0 & 0 & 1  & 2 & 1 & 1 & 0 & $\ 4$\\ \hline
 &  2 & 0 & 1 & 2 & 2 & 1 & 1 & 0 &  $\ 8$\\ \hline
 &  $n$ & 1 & $n$ & 2 & 2 & 1 & 1 & 0 &  $\ 12$\\ \hline
\end{tabular}

Table 3: {\sl The final stack produced by the clausal $n$-algorithm.}

{\bf 4.1} Recall that the premise and conclusion of $(x_6\wedge x_7)\to (x_2\vee x_3\vee x_4\vee x_5)$ triggered an Abraham 0-Flag and Abraham 1-Flag respectively. In general  Abraham 1-Flags have sidelength equal to the number of diagonal 1's. This needs not be true for Abraham 0-Flags since instead of 0 a diagonal entry may also be $n...n$. For instance, imposing the clause $\ol{x}_1\vee\ldots\vee \ol{x}_5$ upon $r'$ in Table 4 yields
$r'_1\uplus r'_2\uplus r'_3$ which features an Abraham 0-Flag  of dimensions $3\times 5$ on the left. (As opposed to Table 2 here only its diagonal elements are rendered boldface.)

\begin{tabular}{l|c|c|c|c|c|c|c|c|c|c|c|c}
& $x_1$ & $x_2$ & $x_3$ & $x_4$ & $x_5$ & & $x_6$ & $x_7$ & $x_8$ & $x_9$ & $x_{10}$ &\\ \hline
$r'=$ & $n_1$  & $n_1$ & $n_2$ & $n_2$ & $n_3$ &    & $n_1$ & $n_2$ & $n_2$& $n_3$ & $n_3$ & \\ \hline
& & & & & & & &  & & & &\\ \hline
$r'_1=$ & ${\bf n_1}$ &${\bf n_1}$ & $n_2$ & $n_2$ & $n_3$ &   & 2 & $n_2$ & $n_2$& $n_3$ & $n_3$ & \\ \hline
$r'_2=$ & 1 & 1 & ${\bf n_2}$ & ${\bf n_2}$ & $n_3$ &   & 0 & 2 & 2 & $n_3$ & $n_3$ & \\ \hline
$r'_3=$ & 1 & 1 & 1 & 1 & {\bf 0} &   & 0 & $n_2$ & $n_2$ & 2 & 2 & \\ \hline
\end{tabular}

Table 4: {\sl A more typical Abraham 0-Flag}

For a complete classification of what can happen when one imposes a (Horn) clause upon a 012n-row,  we refer to [10, Sec.5]. The clausal $n$-algorithm was inspired by, and extends the {\it Horn n-algorithm} of [10, Sec.5]  mainly in two ways. 
First, apart from the Abraham 0-Flags in both algorithms, the clausal $n$-algorithm also  features Abraham 1-Flags (such as the boldface parts in $r_4$ to $r_6$, respectively $r_9$ to $r_{12}$).  
 Second, the clausal $n$-algorithm needs a SAT-solver to check the feasibility of rows since, recall, the feasibility of $r$ amounts to the satisfiability of $T(r) \wedge \varphi$. (In contrast, the Horn $n$-algorithm enjoys a straightforward feasibility test, see 7.1.)

 {\bf 4.2} A few words about the feasibility of rows are in order. Let $r$ be the top row of the working stack. Once the candidate sons of $r$ have taken its place, $r$ can be deleted. This is a {\it harmless} deletion. However if no candidate son of $r$ is feasible, then the deletion (and creation) of $r$ was {\it wasteful} since $r$ contained no $\varphi$-models. Let $wdel$ be the number of wasteful deletions encountered in any fixed run of the clausal $n$-algorithm. The only guarantee to have $wdel=0$ is to employ a feasibility test. Indeed, by induction (anchored in $r= (2, 2, \ldots, 2)$) let $r$ be feasible, i.e. $r$ contains models. Then $r$ has feasible candidate sons. All of them, and only them, are picked by the feasibility test. Yet feasibility testing (i.e. satisfiability testing) is costly, and so it sometimes pays to stick with weak feasibility and accept a moderate amount of wasteful deletions. This is investigated in more depth in other articles.

{\bf 4.3} How does Clausal Imposition compare to Pivotal Decomposition? The biggest benefit of the former is the extra compression provided by the $n$-wildcard. As previously mentioned, the compression tends to be the higher  the longer the clauses are. Another benefit is that 012n-rows are automatically final as soon as the last clause has been imposed. In contrast, Pivotal Decomposition needs a SAT-solver to guarantee the finality of a 012-row, i.e. to guarantee that the accompaning formula (see Table 1) is a tautology. One  advantage of Pivotal Decomposition is that {\it any} formula (such as $\varphi_2$ in (2)) can be digested. If the formula is a CNF, it is allowed to have many clauses, but different from Clausal Imposition not thousands of variables.
(Recall that in the worst case branching upon variables takes time  exponential in the number of variables.)

\section{Fresh air for Min-Cost-Sat and Multiobjective 0-1-programs?}

In 5.1 we first state the Min-Cost-Sat problem. Once the minimum cost (mincost) has been calculated with traditional methods, the clausal $n$-algorithm can take over in order to generate {\it all} mincost solutions. This is driven further towards multiobjective 0-1-programming in 5.2.  The fact that the author only recently got acquainted with existing research [11]  helped to explore the potential benefits of Clausal Imposition more freely.  Some further insight follows in Section 6 but many issues remain to be sorted out.

{\bf 5.1} Given a (component-cost) function $f:\{1,2,...,w\}\to \Z$, and a Boolean function $\varphi:\{0,1\}^w\to \{0,1\}$ in CNF, the {\it Min-Cost-Sat} problem is to find a bitstring $z_{min}\in Mod(\varphi)$ that minimizes the {\it cost} $F(x):=\sum_{i=1}^w f(i) x_i$ where $x:=(x_1,\ldots,x_w)\in \{0,1\}^w$. We put $F^{min}:=F(z_{min})$. For instance, if $f(i):=1$ for all $1\le i\le w$ then $F^{min}$ is the minimum Hamming-weight  of a $\varphi$-model. This special case
of Min-Cost-Sat is known as the {\it Min-Ones} problem.

{\bf 5.1.1} For each $012n$-row $r$ let

 $F^{min}(r):=min\{F(x):\ x\in r\}$ and $Min(r):=\{x\in r: F(x)=F^{min}(r)\}$.

\begin{tabular}{l|c|c|c|c|c|c|c|c|c|c|c|c|c|c|c|}
	$f=$  & $3$  & $4$ & $-5$    & $0$ & $-3$ & $5$ & $8$   &  $-3$ & $0$ & $0$ & $7$ &  $-7$ & $-5$ &  $-5$ & $-5$  \\    \hline
$r=$ & $n_1$ & $n_1$ & $n_1$ & $n_1$ & $2$ & $2$ & ${\bf 1}$    &  $n_2$ & $n_2$ & $n_2$ & ${\bf 0}$ & $n_3$ & $n_3$ &  $n_3$ & $n_3$  \\  \hline
 
$Min(r)=$ & $0$ & $0$ & $1$  & $2$ & $1$ & $0$ & ${\bf 1}$    &  $1$ & $n$ & $n$& ${\bf 0}$ &  $1$ & $g_2$ &  $g_2$ & $g_2$  \\ 

\end{tabular}

{\sl Table 5: How $F^{min}(r)$ depends on the $n$-wildcards of $r$. }

 For instance, for $r$ and $f$ in Table 5 one checks that $F^{min}(r)=-5-3+8-3-7-2\cdot 5=-20$ and that $Min(r)$ is as indicated (the $g$-wildcard of 2.3 reappears here). Hence $|Min(r)|=18$.
It is evident that generally $F^{min}(r)$ and $Min(r)$ can be determined fast. Suppose $F^{min}$ is known, e.g. found by 0,1-programming (more on that in a moment). We can then run the clausal $n$-algorithm as described in Section 4, except that all candidate sons $r$ with $F^{min}(r)> F^{min}$ are discarded right away; the others still need to pass the ordinary (weak or proper) feasibility test. In the end, each final $012n$-row $r_i\ (1\le i\le R)$ contains at least one $x$ of cost $F^{min}$. It follows that $Min(r_1)\uplus\cdots\uplus Min(r_R)$ is the set of {\it all} optimal $\varphi$-models.

{\bf 5.2} Recall that each clause, such as $x_2\vee \ol{x_4} \vee \ol{x_7}$, readily\footnote{Conversely, translating inequalities such as $2x_1-3x_2+6x_3\ge 4\ (x_i\in\{0,1\})$ into Boolean expressions is a well-known harder problem.}
translates to the inequality $x_3+(1-x_4)+(1-x_7)\ge 1$, i.e. to $x_3-x_4-x_7\ge -1$. Hence  $(x_2,x_4,x_7)\in\{0,1\}^3$ satisfies
$x_2\vee \ol{x_4} \vee \ol{x_7}$ iff it satisfies $x_3-x_4-x_7\ge -1$. Thus every Min-Cost-Sat problem translates into a 'CNF-induced' 0-1-program. Many efficient solvers for  0-1-programs exist, but less so for {\it multiobjective} 0-1-programs. In particular, while Mathematica can solve an ordinary 0-1-program, it cannot produce {\it all} its optimal solutions, nor can it handle multiobjective 0-1-programming.
We just showed how Clausal Imposition yields all optimal solutions and now speculate whether it can be applied to multiobjective 0,1-programming as well. We will switch back and forth between a Boolean and a 0-1 programming point of view. In 5.2.1 and 5.2.2 we look at disjunctive and conjunctive objectives respectively.

{\bf 5.2.1} Given are $s\ge 1$ component-cost functions $f_i$ (and their coupled cost functions $F_i$), along with a Boolean function $\varphi:\{0,1\}^w\to\{0,1\}$ in CNF-format.
For fixed $b\in\Z$ we like to settle this 

{\it Question:} Is there an index $\alpha$ and an $x\in Mod(\varphi)$ with $F_{\alpha}(x)\le b$? 

In other words, $x$ must satisfy the {\it disjunctive} objective

 $$F_1(x)\le b\ or\ F_2(x)\le b\ or\ ... \ or \ F_s(x)\le b$$.
 
  As in 5.1 we run the clausal $n$-algorithm but now keep updating the value $\mu$ of the smallest $F_j(y)$ where $1\le j\le s$ and $y$ ranges over the $\varphi$-models output already. Whenever a clause is to be imposed upon the top row $r$ of the working stack, we proceed as follows. If $F^{min}_j(r)>\mu$ for al $1\le j\le s$, then discard $r$. Otherwise calculate the feasible candidate sons $\rho$ of $r$. If some $F_k^{min}(\rho)$ is strictly smaller\footnote{This does not follow merely from our assumption that $F^{min}_j(r)\le\mu$ for some $j$; not even when $\le$ is $<$.} than $\mu$, then update $\mu$ (and the corresponding index $k$) accordingly and discard all other candidate sons. Once the clausal $n$-algorithm has terminated, compare $\mu$ with $b$. If $\mu\le b$, the answer to the Question above is 'yes', otherwise 'no'.

Why not settling the Question by simply running a $0,1$-program for each cost function $F_i\ (1\le i\le s)$? Because our approach,
which presumably is the more competitive the larger $s$, gains information by playing off the $F_i$'s against each other. This is hardly possible when all $F_i$'s are handled individually.

{\bf 5.2.2} Once the index $\alpha$ is found, how do we get {\it all} $\varphi$-models $x$ satisfying $F_\alpha(x)\le b$? Because in 5.2.1 all $012n$-rows that do contain such $x$'s can be set aside, it suffices to show how within {\it one} row $r$ all $x\in r$ with $F_\alpha(x)\le b$ are found. To fix ideas, let $\alpha:=1,\ b:=15$, and let the 012-row $r$ be as in Table 6.

\begin{tabular}{l|c|c|c|c|c|c|c|c|c|c|c|c|c|c}
	& {\bf 1} & 2 & 3 & 4 & {\bf 5}  &  {\bf 6}& 7 & {\bf 8} &  {\bf 9}& &10 & 11 & 12&\\ \hline
	
	$f_1=$ &   {\bf 3} & 3 & 3 & 3 &  {\bf 4} &  {\bf 4} & 4 &  {\bf 5}&  {\bf 5}& & -6
	& 8& 0& \\ \hline
	
	$r=$ & 2  & 2 & 2 & 2 & 2 & 2 & 2 & 2& 2& & 1
	 & 0& 1& \\ \hline

\end{tabular}

{\sl Table 6: Finding all $x\in r$ of bounded cost $F_1(x)$.}

Thus we like to represent, in compact format, all $x\in r$ that satisfy 

$$F_1(x)=f_1(1)x_1+f_1(2)x_2+\cdots +f_1(12)x_{12}\le 15.$$

 Because all these $x$'s have $f_1(10)x_{10}=-6$ and $f_1(11)x_{11}=f_1(12)x_{12}=0$,  it suffices to compress the set
${\cal S}_1$ of bitstrings $y\in\{0,1\}^9$ for which $F_1(y)$ (defined in the obvious way) is $\le 15+6=21$. Conveniently ${\cal S}_1$  is a simplicial complex (i.e. from $y\in {\cal S}_1$ and $y'\le y$ follows $y'\in {\cal S}_1$). We first target its facets (=maximal members) by listing all sums of $f_1$-values which are maximal with respect to being $\le 21$:

$5+5+4+4+3,\ 5+5+4+3+3,\ 5+5+3+3+3,\ 5+4+4+4+3,\\ 5+4+4+3+3,\ 5+4+3+3+3+3,\ 4+4+4+3+3+3,\ 4+4+3+3+3+3$

For instance $5+5+4+4+3$ yields the facet $\{1,5,6,8,9\}$ (boldface in Table 6), but also $\{2,6,7,8,9\}$ and some more.
Likewise, each of the other sums yields several  facets. Knowing the facets of ${\cal S}_1$ one
 can compress the {\it whole} of ${\cal S}_1$ using appropriate\footnote{In fact this is the $e$-wildcard  which is 'dual' to the $n$-wildcard in that it demands 'at least one 1 here'.
 The $e$-wildcard is exploited in the Facets-To-Faces algorithm (arXiv:1812.02570) which represents any simplicial complex given by its facets as a disjoint union of $012e$-rows. The $e$-wildcard also compresses the collection of all spanning trees of a graph (arXiv:2002.09707), the collection of all minimal hitting sets of a hypergraph (arXiv:2008.089960), and it boosts the calculation of expected trial lengths in Coupon Collector problems [12].} wildcards.

{\bf 5.2.3} As if one cost function $F_1$ wasn't enough, let us consider $s$ different cost functions $F_i$ and ponder the {\it conjunctive} objective to find the set ${\cal S}$ of all $x\in r$ satisfying

$$F_1(x)\le b_1\ and\ F_2(x)\le b_2\ and\  .....\ and\ F_s(x)\le b_s.$$

Evidently ${\cal S}={\cal S}_1\cap\ldots \cap {\cal S}_s$ where ${\cal S}_i$ is the simplicial complex triggered by $F_i$ (see 5.2.2). Pleasantly the facets of ${\cal S}$ are smoothly obtained from the facets of the ${\cal S}_i$'s. They are just the inclusion-maximal members among the sets $A\cap B\cap\ldots\cap C$, where $A,B,\ldots,C$ range, respectively, over the facets of
${\cal S}_1,{\cal S}_2,\ldots ,{\cal S}_s$.  Notice that the mere cardinality of ${\cal S}$  can be calculated faster. In particular, the conjunctive objective has no solution iff $|{\cal S}|=0$.

 We mention that matters get more complicated when proper $012n$-rows are considered (as opposed to the 012-row in Table 6).

\section{ Numerical experiments}

 The hardwired Mathematica command {\tt BooleanConvert} can calculate (among other options) an ESOP for a given Boolean function $\varphi$. The second Mathematica player is  {\tt SatisfiabilityCount} which calculates $BDD(\varphi)$ and from it $|Mod(\varphi)|$. Unfortunately the user has no access to $BDD(\varphi)$  and thus cannot calculate  the BDD-induced ESOP (see 2.2). On whatever techniques {\tt BooleanConvert} is based, it is {\it not} BDD's (as inquired by the author). This leads one to speculate that {\tt BooleanConvert} usually achieves a better compression than the BDD-induced ESOP\footnote{This should be interesting news to Toda and Soh who in [13] advertise the use of BDD's to compactly enumerate all models of $\varphi$.}, notwithstanding the fact (witnessed by the times of {\tt SatisfiabilityCount}) that the BDD is calculated faster.
  In view of Subsection 2.5 we may further speculate that {\tt BooleanCovert} is closely related to Pivotal Decomposition.
  
   How {\tt BooleanConvert}  fares against the clausal $n$-algorithm, both time-wise and compression-wise, will be discussed in Subsections 6.1.1 to 6.1.5. All examples  are such that the clausal $n$-algorithm fares better by omitting feasibility tests (see 4.2).
In 6.2 and 6.3  not a competition but  a collaboration in the spirit of Section 5 between Clausal Imposition and {\tt LinearProgramming} (the third Mathematica player) takes place.

{\bf 6.1.1} Jumping into medias res, the first numerical line in Table 7 means the following. A CNF with $w=50$ variables and $\lambda=25$ clauses was built, each of which consisting of $neg=10$ random negative  literals.   It took {\tt SatisfiabilityCount} 0.9 seconds to find the precise  number of models. The clausal $n$-algorithm packed the models into 315'833 many $012n$-rows $r$. On average $r$ featured (rounded) 15 don't-care 2's and 8 $n$-symbols, distributed among 3 $n$-bubbles.
The running time was 94 seconds. The {\tt BooleanConvert} command could not handle the task and automatically aborted. 
The (50,25,10,2)-instance has clauses with 10 negative and 2 positive literals. By whatever reason, {\tt BooleanConvert} could handle it in short time, yet its compression trails behind Clausal Imposition. Similar remarks apply to the (50,25,10,5) and (50,25,10,10) instances.

{\bf 6.1.2} If we keep $h=10$ small but increase $w=50$ to $w=2000$ then  {\tt BooleanConvert} suffers compression-wise {\it and} time-wise. In extreme cases such as (30000,8,29000,0) even \\ {\tt SatisfiabilityCount} stumbles; the 453 seconds 'timed' by Mathematica in reality were several hours. (Could it be that some massive data transfers were simply not timed?)

\begin{tabular}{|l|c|l|l|} \hline 
  &     &   & \\
\ $w \ \ \ \lambda\ \ neg\ \ pos$ & {\tt SatCount} & \ \ \ \ \ \ \ \ Clausal $n$-algorithm & {\tt BooleanConvert} \\
    &   Time     & 012n-rows\ \ \ \ \ \ \ \ \ \ \ \ \ \ \ \ \ \ \ Time \ \  & 012-rows \ \ \ Time \ \     \\ \hline
		
50\ \ 25\ \ \ 10\ \ \ 0 & 0.9   & 315'883\ (15,8,3)\ \ \ \ \ \ \ \ \ \ 94      & self-aborted\\ \hline
50\ \ 25\ \ \  10\ \ \ 2 & 0.1   & 378'733\ (15,5,2)\ \ \ \ \ \ \ \ \ \ 119      & 1'270'530\ \ \ \   \ \ 8\\ \hline
50\ \ 25\ \ \ 10\ \ \ 5 & 0   & 21'816\ (18,5,2)\ \ \ \ \ \ \ \ \ \ \ \ 6      & 224'947\ \ \ \ \ \ \ 0.9\\ \hline
50\ \ 25\ \ \ 10\ \ \ 10 & 0   & 7239\ (23,4,1)\ \ \ \ \ \ \ \ \ \ \ \ \ \ 2      & 17'982\ \ \ \ \ \  \ \ \ 1\\ \hline
50\ 100\ \ 10\ \ \ 10 & 0.1   & 319'828\ (19,4,1)\ \ \ \ \ \ \ \ \ \ \  233      & 305'007\ \ \ \ \ \   \ \ 3\\ \hline
 
 &  & & \\ \hline
2000\ 10\ \ \ 800\ 0 & 3 & 110'283\ (490,282,5)\ \ \ \ \ 155 & self-aborted \\ \hline
2000\ 10\ \ \ \  800\ 30 & 0.4 & 1143\ (1150,58,1)\ \ \ \ \ \ \ \ \ \ 2 & 260'731\ \  \ \ \ \ 14\\ \hline
2000\ 1000\ 800\ 30 & $>24$ hours & tops {\tt SatisfiabiliyInstances}\ \ \ \ \ \  & self-aborted\ \  \ \ \ \   \\ \hline
     &  &  & \\ \hline
30000\ 8\ 29000\ 0 & claimed 453   & 140\ (132,1107,2)\ \ \ \ \ \ \ \ \ \ \ 9      & self-aborted \\ \hline
&  & & \\ \hline
50\ \ \ 14\ \ \ \  3\ \ \ 2(2)  & 0.1   & 113'334 (21,4,2)\ \ \ \ \ \ \ \ \ \ \ \  45      & 4'193'743\ \ \ \ \ \  43 \\ \hline
50\ \ \ 15\ \ \ \ 3\ \ \ 2(2)  & 0.3   & 883'424 (18,4,2)\ \ \ \ \ \ \ \ \ \ \ \    363      & self-aborted after 17 min\ \ \ \ \ \   \\ \hline
50\ \ \ 25\ \ \ \ 3\ \ \ 2(2)  & 29   & \ \ \ \ \ \ \ \ \ \ \ \ \ \ \ extrapolated 12 hours      &  self-aborted\ \ \ \ \ \   \\ \hline
50\ \ \ 5\ \ \ \ \ 5\ \ \ 5(5)  & 0.1   & 26'156\ (12,5,2)\ \ \ \ \ \ \ \ \ \ \  \ \ \ 7      & 3'270'328\ \ \ \ \ \   26\\ \hline
50\ \ \ 6\ \ \ \ \ 5\ \ \ 5(5)  & 0.8   & 83'228\ (14,4,2)\ \ \ \ \ \ \ \ \ \ \  \ \ \ 27      &\ \  \ \ \ \ \ \ \ \ \ \ \ \   $>1$ hour\\ \hline
 \end{tabular}

 {\sl Table 7: Clausal Imposition versus BooleanConvert.}

{\bf 6.1.3} For a random $(2000, 1000, 800, 30)$ instance neither {\tt BooleanConvert}, nor {\tt Satisfiability}
{\tt Count}, nor the clausal $n$-algorithm could complete the  task within 24 hours. However, at least the latter produced something. And so did {\tt SatisfiabilityInstances} (option `TREE'). But while {\tt SatisfiabilityInstances} generated a few million  instances (=models)
 one-by-one, the clausal $n$-algorithm  produced more than $3$ millions {\it final  rows}, each one of which containing zillions of instances. Thus for certain types of problems, the clausal $n$-algorithm tops {\tt SatisfiabilityInstances} as a generator of random\footnote{Whether `random' in the proper sense of the word applies to {\tt SatisfiabilityInstances} cannot be decided by a layman user. As to the clausal $n$-algorithm, final rows (and hence models) {\it can} be generated at random. This is easily achieved by frequently permuting the working stack (Section 4). In this way also the achieved compression and total running time can be extrapolated without Clausal Imposition having to terminate.} models.

{\bf 6.1.4} Both {\tt BooleanConvert} and {\tt SatisfiabilityCount} are hardwired Mathematica commands which hence have a time-advantage over the high-level Mathematica encoding of the clausal $n$-algorithm. One could ponder how fast a hard-wired clausal $n$-algorithm might be. But more importantly, Clausal Imposition is amenable to distributed computation  (aka parallelization).  Indeed,  the 012n-rows in the working stack of Section 4 are completely independent of each other, and can hence at any stage be distributed to distinct processors. To put it bluntly, all it takes to speed up Clausal Imposition by a factor 1000, is to find 1000 colleagues with the same (or better) desktop willing to participate. To the author's best knowledge, the calculation of  a BDD (and hence  {\tt SatisfiabilityCount}) cannot be distributed. As to {\tt BooleanConvert}, only its programmer can tell.

\newpage

{\bf 6.1.5} What  if the number $h$ of clauses is larger\footnote{One referee remarked that this in fact is more relevant in practise. While this may be the case, it is easy to come up with instances of the other kind, e.g. in health care  (or genetics): One may wish to classify the thousands (= $w$) of citizens (proteins) of a city (organism) according to $h=150$ medical, social or other criteria.}  than the number $w$ of variables? Then {\tt BooleanConvert} wins out, always time-wise, sometimes also compression-wise. We content ourselves with (50,100,10,10) as the only instance of that kind. Perhaps combining Clausal Imposition with the pivot selection strategy of 2.5.2.2 (in order to reduce the number of clauses still to be imposed) would save the day. It is clear that a combination of the two is easy to program, but this has not been carried out yet. In contrast, promising numerical experiments in other contexts are discussed in 
 6.2 and 6.3 (and later 7.3).

{\bf 6.2} In Table 8 we keep on evaluating random CNF's  ${\varphi}$ of type $(w,\lambda,neg,pos)$. As described in Section 5 these CNF's are then translated into 0-1-programs and accompanied by a random component-cost function $f:\{1,...,w\}\to [c,d]$. Here $[c,d]:=\{i\in \Z:\ c\le i\le d\}$. The smaller the interval $[c,d]$ the more optimal $\varphi$-models tend to occur. The Mathematica command {\tt LinearProgramming} (always option '0-1') calculates  {\it one} optimal ${\varphi}$-model $z_{min}$, i.e. $F(z_{min})=F^{min}$.   Upon receiving $F^{min}$ the adapted clausal $n$-algorithm calculates {\it all} optima (see 5.1). We record their number as well as the number of $012n$-rows housing them; the average of five trials is taken here. While the clausal $n$-algorithm struggled with the (50,100,10,10)-instance in Table 7, in the present context even a (50,500,10,10)-instance goes down well (Table 8).

\begin{tabular}{|l|c|l|c} \hline 
  &     &   & \\
\ $w \ \ \ \lambda\ \ \ \ neg\ \ pos\ \ \ \ \ [c,d]$ & {\tt LinearProgramming} &\ \ \ \ \ \ \ \  Clausal $n$-algorithm   \\
    &   Time     & optimal sol.\ \ \ \ 012n-rows\ \ \ \ Time \ \  \      \\ \hline
		
50\ \ 500\ \ 10\ \ \ \ \ 10\ \ \ [-20,20] & 0   & 2.2\ \ \ \ \ \ \ \ \ \  \ \ \ \ \  \ (1.4)\ \ \ \ \ \ \ \ \ \ \ \ \ 0.1    \\ \hline
50\ \ 500\ \ 10\ \ \ \ \ 10\ \ \ \ \ [-4,4] & 0   & 42\ \ \ \ \ \ \ \ \ \ \ \  \ \ \ \ \ \ (6)\ \ \ \ \ \ \ \ \ \ \ \ \ 0.2    \\ \hline
200\ \ 1000\ \ 800\ \ \ 30\ \ \ [-4,4] & 0.1   & $\approx 10^{74}$\ \ \ \ \ \ \ \ \ \ \  \ (3.7)\ \ \ \ \ \ \ \ \ \ \ \ \ 5    \\ \hline
 \end{tabular}

{\sl Table 8: Computing {\bf all} optima of a 0-1 program.}

Let us quickly compare with BDD's. While in [7,p.209] Knuth shows how to find one (not all) mincost model from the BDD of $\varphi$, one needs the {\it whole} BDD to do so. This is much different to Clausal Imposition which at an early stage discards infeasible candidate sons.

{\bf 6.3} For a random instance of type $(w,\lambda,neg,pos)=(200,500,5,5)$ with coupled random component-cost functions $f,\ g,\ h$ (having range in [-20,20]) we used {\tt LinearProgramming}  to calculate the corresponding values $F^{min},\ G^{min},\ H^{min}$ (each took about 0.015 seconds). See Table 9 where we also list the times it took to follow up with the clausal $n$-algorithm to find all optima in each case.

\begin{tabular}{c|c|c}
  Weight function  &  Minimum weight  & Time for all optima \\ \hline 
 $f$  &  $F^{min}=-1108$ & 1.6 sec \\ \hline 
 $g$  &  $G^{min}=-947$ & 4.2 sec \\ \hline 
 $h$  &  $H^{min}=-1034$ & 5.4 sec \\ \hline 
\end{tabular}

{\sl Table 9: Venturing into multiobjective 0-1 programming.}

What about optimizing $F,G,H$ simultaneously? What does that really mean? For instance, finding an $x\in\{0,1\}^{200}$ that minimizes $F(x)+G(x)+H(x)$ merely amounts to minimizing {\it one} other  function (i.e. $k:=f+g+h$ which predictably has $K_{min}\ge F_{min}+G_{min}+H_{min}$), and so one doesn't leave the realm of ordinary 0-1 programming. Yet  other questions come to mind, such as: Is there an $x\in\{0,1\}^{200}$ having $F(x),\ G(x),\ H(x)\le -947$? As discussed in 5.2.3, the clausal $n$-algorithm answers 'no' in 0.01 seconds. If -947 is replaced by -870, the aswer is still 'no' but takes 45 seconds. For -500 the answer is 'yes' and takes 315 seconds. Recall  that 'yes' has the potential to be refined to a compressed representation of {\it all} feasible instances. Collaboration on these issues  is welcome.

\section{Variations (past and future) of Clausal Imposition}

In Subsection 7.1 and its interleaved sub-subsections we discuss Horn CNFs. Subsection 7.2 is about 2-CNFs. Many of the algorithms presented in 7.1 and 7.2 have been implemented (and some published) a while ago. Although the clausal $n$-algorithm is fit to handle their tasks, these old implementations are taylor-made and hence potentially faster for their specific inputs (yet this has not been investigated).
Instead of specializing clauses,
in 7.3 we {\it generalize} them  to superclauses. 

 For the sake of unraveling hidden relationships the author may be forgiven for citing several of his own articles/preprints in this Section.

{\bf 7.1} By definition a {\it Horn} clause features at most one positive literal. Any CNF consisting of Horn clauses is called a {\it Horn CNF}.  
As mentioned in 4.1, the clausal $n$-algorithm was inspired by the Horn $n$-algorithm. When $R$ is the number of final 012n-rows, the latter has [10,Thm 2] a running time of $O(R^2h^2w^2)$. (This will be rederived from a higher vantage point in Section 10.) While the Horn $n$-algorithm formally runs in polynomial total time, i.e. does not boast polynomial delay like [9], in practise it is often superior due to its compressed output.

{\bf 7.1.1} A Horn clause is {\it proper} (or {\it pure}) if it features a positive literal, such as $\ol{x_7}\vee \ol{x_8}\vee x_4$. This proper Horn clause  is equivalent to the formula $(x_7\wedge x_8)\to x_4$. If we have a couple of proper Horn clauses with the {\it same} premise, say $(x_7\wedge x_8)\to x_4$ and $(x_7\wedge x_8)\to x_5$
and $(x_7\wedge x_8)\to x_6$ then their conjunction is equivalent to $(x_7\wedge x_8)\to (x_4\wedge x_5\wedge x_6)$. It follows that each proper Horn CNF is equivalent to a conjunction of these kind of {\it implications}\footnote{Talking about {\it implications} such as $\{7,8\}\to\{4,5,6\}$ (which is smoother 
than $(x_7\wedge x_8)\to (x_4\wedge x_5\wedge x_6)$	) started in Formal Concept Analysis in the late 80's. In other Data Mining scenarios one speaks of {\it association rules} 
or {\it functional dependencies}. We emphasize the major difference between implications and {\it clausal} implications such as $(x_7\wedge x_8)\to (x_4\vee x_5\vee x_6)$ (see Section 4).}. 

 {\bf 7.1.1.1:} Consider the specific case where all implications have singleton premises, i.e. are of type $x_i\to (x_s\wedge x_t\wedge..)$. Such an implication, and whence the whole proper Horn CNF, is equivalent to a conjunction of implications $x_i\to x_s,\ x_i\to x_t$, and so forth. Take these implications as the directed edges of a graph. Factoring out its strong components yields a poset which is uniquely determined by listing the lower covers $b_1,\ b_2,\ ...$ of each (non-minimal) element $a$. This gives rise to the wildcard $(a,b,\ldots,b)$ and a corresponding $(a,b)$-algorithm (Order 31 (2014) 121-135) that yields all order ideals of a poset in a compressed format. 
 
{\bf 7.1.1.2} It is work in process that implications, all of whose premises have cardinality {\it two}, can be condensed in similar ways. For instance, the subgroup-lattice of a group $(G,\star)$ can be calculated in compressed format by imposing all implications of type $\{a,b\}\to\{a\star b\}
$. Similar for semigroups or in fact any universal algebra with merely binary (and unary) operations.

{\bf 7.1.2} What is the best way to think of a  Horn CNF $\varphi$ which consists entirely of {\it improper} Horn clauses
(i.e. having exclusively negative literals)?
Identify the set of bitstrings $\{0,1\}^w$ with the powerset ${\cal P}[w]$ of $\{1,2,..,w\}$ in the usual way and let $C_1,...,C_h\in {\cal P}[w]$  match the  improper Horn clauses. Call $X\in {\cal P}[w]$ a {\it noncover} if $C_i\not\subseteq X$ for all $1\le i\le h$. Then $\varphi(X)=1$ iff $X$ is a noncover. 
	In order to enumerate $Mod(\varphi)$ a simplification of the Horn $n$-algorithm can be used. This {\it noncover n-algorithm} [10] has a trivial feasibility test (as opposed to the more subtle $O(hw)$ feasibility test of the general Horn $n$-algorithm).

{\bf 7.1.2.1} Consider the special case where $C_1,...,C_h\in {\cal P}[w]$ are the edges of a graph $G$ with vertex set $V=\{1,...,w\}$. Then the noncovers with respect to $C_1,...,C_h$ are exactly the anticliques (=independent sets) of $G$. It pays
 to clump edges that share a common vertex. In formulas, $(\ol{x}_1\vee\ol{x}_2)\wedge (\ol{x}_1\vee\ol{x}_3)\wedge (\ol{x}_1\vee\ol{x}_4)$ is equivalent to $x_1\to (\ol{x}_2\wedge\ol{x}_3\wedge\ol{x}_4)$. Thus if vertex 1 is a member of a 'wannabe' anticlique $X$ then its neighbours 2,3,4 must not be in $X$. By definition the set of nine bitstrings $(x_1,x_2,x_3,x_4)\in\{0,1\}^4$ satisfying $x_1\to (\ol{x}_2\wedge\ol{x}_3\wedge\ol{x}_4)$ is denoted by $(a,c,c,c)$. This wildcard gives rise to the $(a,c)$-algorithm that enumerates in compressed format all anticliques of a graph (arXiv:0901.4417).

{\bf 7.2} CNF's with clauses having at most two literals are called 2-CNF's. Provided the 2-CNF $\varphi$ is satisfiable (which can be tested fast) one can rename [5, p.236] the literals of $\varphi$ in such a way that the resulting formula $\psi$ is a  {\it Horn} CNF, and such that there is a straightforward bijection between the models of $\varphi$ and $\psi$. As seen, the models of $\psi$ can be computed in polynomial total time and in a compressed format. Furthermore, since $\psi$ has only 2-clauses,  the Horn $n$-algorithm can be accelerated by combining the $(a,c)$-algorithm of 7.1.2.1 with the $(a,b)$-algorithm of 7.1.1.1. This is work in progress (arXiv:1208.2559).

{\bf 7.3} While  7.1 and 7.2 were about specializations of the clausal $n$-algorithm, we now turn to a {\it generalization} of it. Namely, instead of normal clauses like $\ol{x}_1\vee\ol{x}_2\vee x_3\vee x_4\vee x_5$ we look at {\it superclauses} like

(7)\quad $\ol{x}_1\vee\ol{x}_2\ \vee\ (x_3\wedge x_6)\vee (x_4\wedge x_2\wedge x_5)\vee (x_5\wedge x_3\wedge x_1)$.

Thus each positive literal can give way to a conjunction of positive literals (call that a {\it pos-term}). Any conjunction of superclauses will be called a {\it super-CNF}. Each superclause  can of course be turned into a conjunction of normal clauses, but this is labor-intensive and can  trigger thousands of clauses.
 Some combinatorial problems are naturally phrased in terms of superclauses and therefore the author has implemented the imposition of 
superclauses 'head-on' in high-level Mathematica code and calls the result {\it superclausal n-algorithm}.

\newpage 

As announced in 6.1.5, let us rekindle the battle between Clausal Imposition and {\tt BooleanConvert} 
 in the context of super-CNF's. Thus in Table 7 of Section 6 the instance (50,14,3,2(2)) signifies that we randomly generated a super-CNF on 50 variables consisting of 14 superclauses all of which featuring 3 negative literals and 2 pos-terms (each  of length 2). The five instances show that Clausal Imposition benefits more from super-CNF's than {\tt BooleanConvert}. As opposed to the former, for the latter it can pay to first switch to ordinary CNF, and then run {\tt BooleanConvert}. That depends on the size of the CNF which variates strongly\footnote{The CNF equivalent to the (50,5,5,5(5))) super-CNF had over 2 million clauses, whereas the one equivalent to the (50,14,3,2(2)) super-CNF had only 56 clauses.} in our examples.

\section{Abstract row-splitting mechanisms}

We introduce a framework  for compressing $Mod(\varphi)$ that covers both Pivotal Decomposition and Clausal Iimposition (and perhaps  other  methods in spe).

{\bf 8.1} Fix a Boolean function $\varphi$ of 'arity' $w=|\varphi|$, and let the {\it valency} $h=h(\varphi)$ be any positive integer. Let $SpMod(\varphi)\subseteq Mod(\varphi)\subseteq {\cal P}[w]$ be any subset the members of which we call {\it special} models. Let ${\cal R} \subseteq {\cal P} ({\cal P}[w])$ be such that ${\cal R}$ contains the powerset ${\cal P}[w]=(2,2,\ldots,2)$. Each $r\in {\cal R}$ is called a {\it row}. Akin to Section 3 call $r\in{\cal R}$ 
 {\it feasible} if $r\cap SpMod(\varphi)\not =\emptyset$, and  {\it final} if $r\subseteq SpMod(\varphi)$. Let $deg:\N\to {\cal R}$ be a {\it degree} function, i.e. it ties in with the concepts above in the sense that for all $r\in{\cal R}$ it follows from $deg(r)=h$ that $r$ is final. Under these circumstances the triplet $({\cal R},deg,h)$ is a {\it row-structure} of $SpMod(\varphi)$.
 
 Before we continue with more terminology in 8.2, a word on $SpMod(\varphi)$ is in order. To unclutter matters the reader is advised to imagine $SpMod(\varphi)=Mod(\varphi)$ throughout Section 8. Throughout Section 9 we will have $SpMod(\varphi)=Mod(\varphi,k)$, i.e. the set of models of Hamming-weight $k$. Whether other choices of $SpMod(\varphi)$  will ever be fruitful remains to be seen.

{\bf 8.2} Given a row-structure of $SpMod(\varphi)$, suppose there is an algorithm ${\cal A}_1$ that for each $r\in{\cal R}$ calculates $deg(r)$ in time  at most $d=d(\varphi)$. Further there is an algorithm ${\cal A}_2$ which for each feasible but non-final $r\in {\cal R}$ achieves the following in time at most\footnote{It will be convenient to further postulate $s\ge w$.}  $s=s(\varphi)$. It finds $\tau\ge 1$ many $r_i\in{\cal R}$ such that:

(a) all  $r_i$ are feasible;

(b) all  $r_i$ have deg$(r_i) > \mbox{deg}(r)$;

(c) $ r \cap Sp\mbox{Mod}(\varphi)=(r_1 \uplus \cdots \uplus r_\tau) \cap Sp\mbox{Mod}(\varphi) $.

We  call the rows $r_i$ the {\it sons} of $r$ and call the transition from $r$ to $r_1,\ldots,r_\tau$ a {\it row-splitting}. Finally the quadruplet 
$({\cal A}_1,d,{\cal A}_2,s)$ is a {\it row-splitting mechanism} for the row-structure $({\cal R},deg,h)$
of $Sp\mbox{Mod}(\varphi) $.
 In practise one may need a SAT-solver as a subroutine of ${\cal A}_2$ yet SAT-solvers do not enter the formal definition of a row-splitting mechanism.

\begin{tabular}{|l|} \hline\\
{\bf Theorem }: Suppose $({\cal A}_1,d,{\cal A}_2,s)$ is a  row-splitting mechanism  for the 
row-structure \\ 
$({\cal R},deg,h)$ of $SpMod(\varphi)$.  If $Sp\mbox{Mod}(\varphi)\neq \emptyset$ then $Sp\mbox{Mod}(\varphi)$ \\
can be enumerated, using $R$ many disjoint rows, in total time $O(Rh(d+s))$.\\
\\
\hline \end{tabular}

{\it Proof:} Put $w:=|\varphi|$. Because $Sp\mbox{Mod}\not=\emptyset$ the row $r_0:={\cal P}[w]$ is feasible, and so by (a) there is a tree
${\cal T}_1$ with root $r_0$ and leaves $r_1,\ldots, r_\tau$ satisfying (c). Being true for ${\cal T}_1$ assume by induction that we obtained\footnote{In computational practise a LIFO stack incorporates the tree ${\cal T}_1$ .} a tree ${\cal T}_i$ with root $r_0$ and  feasible leaves $\rho_1,\ldots, \rho_k$ whose union is disjoint and contains $Sp\mbox{Mod}(\varphi)$.

{\it Case 1:} ${\cal T}_i$ has non-final leaves. Then pick the first such leaf, say $\rho_j$, and split it into sons $r_1,\ldots, r_\tau$. 
This makes $\rho_j$ an interior node of a new tree ${\cal T}_{i+1}$ whose leaves are feasible and can be  ordered (say) as 
$\rho_1,\ldots,\rho_{j-1},r_1,\ldots, r_\tau,\rho_{j+1},\ldots,\rho_k.$ Since by assumption
$\rho_1\cup\cdots\cup\rho_j\cup\cdots\cup\rho_k \supseteq Sp\mbox{Mod}(\varphi),$ 
and since by (c) all models contained in $\rho_j$ are contained in
$r_1\uplus\ldots\uplus r_\tau$, and since $\rho_j\supseteq r_1\uplus\cdots \uplus r_\tau$, it follows that

(8)\quad $\rho_1\uplus\cdots\uplus\rho_{j-1}\uplus r_1\uplus\ldots\uplus r_\tau\uplus\rho_{j+1}\uplus\cdots\rho_k\supseteq Sp\mbox{Mod}(\varphi)$.

{\it Case 2:} All leaves $\rho_1,\ldots ,\rho_k$ of
 ${\cal T}:={\cal T}_i$  are final. Then $\rho_1\uplus\cdots\uplus \rho_k\subseteq Sp\mbox{Mod}(\varphi)$. Since "$\supseteq$" always
 holds by (8), equality takes place. Notice that Case 2 eventually {\it does occur} because by (b) every row-splitting strictly increases the degrees of the sons, and sons of degree $h$ are final by definition of a $\varphi$-row-structure.

As to the cost analysis, since there are $|{\cal T}| - R$ many branching nodes, and they are bijective to the  occured row-splittings, the cost of the latter amounts to $O(s(|{\cal T}| - R)) = O(s|{\cal T}|)$. Because the height of ${\cal T}$ is $\leq h$ we get $|{\cal T}| \leq Rh$. By the above the total cost of calculating degrees is $O(Rhd)$. Furthermore, stacking or outputting a (final) length $w$ bitstring costs $O(w)$. Hence, and because we postulated $w \leq s$, the overall cost is
$O(Rhd)+O(Rh s) + O(Rhw) = O(Rh (d+s)).$ \hfill $\square$

All that can be {\it proven} about the number $R$ of rows is that $0 < R \leq |Sp\mbox{Mod}(\varphi)|$. Here $<$ holds because 
 $Sp\mbox{Mod}(\varphi) \neq \emptyset$ and $\leq$ is due to the disjointness of rows.  In practise (Section 6) often $R$ is much smaller than
$|Sp\mbox{Mod}(\varphi)|$.

\section{Pivotal Decomposition  as a row-splitting mechanism}

We reviewed Pivotal Decomposition in 2.5 with a toy example. In Section 9 the level is more abstract; this has pros such as Corollary 1 below, and cons such as abandoning on-the-fly branching in Table 1.
We  first show (9.1) how Pivotal Decomposition can be viewed as a  row-splitting mechanism. Then the three results in 9.2 follow smoothly as Corollaries of the `Master Theorem' in Section 8.
Specifically,
 Corollary 1 shows how a given ESOP of $Mod(\varphi^c)$ induces an ESOP of $Mod(\varphi)$. Corollary 2 is technical and prepares the ground for Corollary 3 which is about $SpMod(\varphi)=Mod(\varphi,k)$, i.e. the set of $\varphi$-models of fixed Hamming-weight $k$.

{\bf 9.1} Throughout Section 9 in all row-structures $({\cal R},deg,h)$  of $SpMod(\varphi)$ the set ${\cal R}$ comprises all  012-rows of length $w$, and deg$(r) : = \mbox{min}(\mbox{twos}(r)) -1$ is the longest 01-prefix of $r$.
 For instance $r = (0,1,1,0, 1, 2, 1, 0, 2, 1)$ has deg$(r)  = 5$, and deg$((2,2,2))=0$.
  Suppose $r$ is feasible and $q: = \mbox{deg}(r) < w$. Let $\rho_0$ and $\rho_1$ be the rows arising from\footnote{We mention in passing that by induction only $012$-rows of type $(\ast, \cdots, \ast, 2, \cdots, 2)$ with $\ast \in \{0,1\}$ will ever be subject to row-splitting.} $r$ by substituting the 2 at position $q+1$ by 0 and 1 respectively.
	Since $r$ is feasible and $r=\rho_1\uplus\rho_2$, not both $\rho_1$ and $\rho_2$ can be infeasible. The one or two feasible rows among them by definition are the sons of $r$ (called $r_1$, or $r_1,\ r_2$). Hence (a), (b), (c) are clearly satisfied. We see that the time $s=s(\varphi)$ to achieve a row-splitting amounts (upon replacing it by $O(s(\varphi))$) to the time for (a), which in turn  boils down to the {\it time for testing the feasibility} of a length $w$ row $\rho$. In particular, $\rho=(2,2,\ldots, 2)$ is feasible iff $SpMod(\varphi)$ is nonempty.
	
{\bf 9.2} By the Master Theorem in Section 8  a nonempty $Mod(\varphi)$ can be enumerated as disjoint union of $R$ rows in time $O(Rh(d+s))$.
If the row-splitting mechanism models Pivotal Decomposition, then
by 9.1 the latter becomes $O(Rw(w+s))$, which is $O(Rws)$ in view of $w\le s$. One can drop the condition that $Mod(\varphi)$ be nonempty by switching from $O(Rws)$ to $O((R+1)ws)$. We call the latter the {\it PivDec-bound}.

 \begin{tabular}{|l|} \hline \\
{\bf Corollary 1:} Suppose that for the Boolean function $\varphi : \{0,1\}^w \ra \{0,1\}$ an ESOP of\\
$\mbox{Mod}(\varphi^c)$ is known which uses $t$ many disjoint $012$-rows. Then $\mbox{Mod}(\varphi)$ can be\\
enumerated, using $R$ many disjoint $012$-rows, in time $O((R+1)tw^2)$. \\
\\
\hline \end{tabular}

{\it Proof:} 
Suppose that our given ESOP is $\mbox{Mod}(\varphi^c) = r'_1 \uplus \cdots \uplus r'_t$. If we can show that the $\varphi$-feasibility of rows $\rho$ can be decided in time $s(\varphi)=O(wt)$ then the PivDec-bound $O((R+1)ws(\varphi))$ becomes $O((R+1)tw^2)$ as claimed.
The $\varphi$-feasiblity of $\rho$  is equivalent to $\rho \cap \mbox{Mod}(\varphi) \neq \emptyset$, which amounts to $\rho \not\subseteq \mbox{Mod}(\varphi^c)$, which amounts to $|\rho \cap \mbox{Mod}(\varphi^c)|< |\rho|$. This inequality can be tested as follows. First $|\rho| = 2^\gamma$ where $\gamma : = |\mbox{twos}(\rho)|$. Second

 $\rho \cap \mbox{Mod}(\varphi^c) = (\rho \cap r'_1) \uplus (\rho \cap r'_2) \uplus \cdots \uplus (\rho \cap r'_t).$

If zeros$(\rho) \cap \mbox{ones}(r'_i) \neq \emptyset$ or $\mbox{ones}(\rho) \cap \mbox{zeros}(r'_i) \neq \emptyset$ then $\rho \cap r'_i = \emptyset$. Otherwise, as seen in 3.2.1, $\rho \cap r'_i$ can again be written as a $012$-row. Hence $|\rho \cap \mbox{Mod}(\varphi^c)|$ can be calculated in $O(wt)$ time. 
 $\square$

A notable special case of $\rho$ being feasible, i.e. satisfying $|\rho \cap \mbox{Mod}(\varphi^c)| < | \rho|$, is that $|\rho \cap \mbox{Mod}(\varphi^c)| =0$. This amounts to $\rho \subseteq \mbox{Mod}(\varphi)$, i.e. to the finality of $\rho$. The enumeration of Mod$(\varphi)$ can thus entail proper $012$-rows $\rho$.
This illustrates that in the definition of a row-structure the condition $deg(\rho)=h$ is sufficient, yet not necessary for $\rho$ to be final. Notice that instead of Pivotal Decomposition  one could also use Clausal Imposition to obtain $Mod(\varphi)$ as a disjoint union of 012n-rows in polynomial total time. This is because the intersection of a 012n candidate son $\rho$ with a 012-row from $Mod(\varphi^c)$ is again a 012n-row, whose cardinality is readily determined.

{\bf 9.2.1}  While the satisfiability of $\varphi$ (Is $SpMod(\varphi)=\emptyset$?) must not be confused with the feasibility of a row $r$ 
(Is $SpMod(\varphi)\cap r=\emptyset$?), in Corollaries 2 and 3 we reduce the latter to the former by virtue of 'heritance'.
Thus call a class ${\cal C}$ of Boolean functions {\it hereditary} if for each $\varphi \in {\cal C}$ the substitution of variables with 0 or 1 yields again an element $\psi$ of ${\cal C}$: We then write $\psi\le\varphi$. Call a function $sat:{\cal R}\to\N$ {\it monotone} if
for all $\varphi\in{\cal C}$ it follows from $\psi\le\varphi$ that $sat(\psi)\le sat(\varphi)$.

\begin{tabular}{|l|} \hline \\
{\bf Corollary 2:} Let ${\cal C}$ be a hereditary class of Boolean functions and let $sat:\ {\cal R}\to\N$  \\
 be monotone. Suppose that for each $\psi\in{\cal C}$ it can be tested in time \\
 $\le sat(\psi)$ whether or not $SpMod(\psi)=\emptyset$. Then for each $\varphi \in {\cal C}$ of arity $w$ one  \\
  can enumerate $SpMod(\varphi)$ in $O((R+1)wsat(\varphi))$ time where $R=|SpMod(\varphi)|$.\\
\hline \end{tabular}

{\it Proof:} The PivDec-bound states it costs $O((R+1)ws(\varphi)$ time to enumerate $SpMod(\varphi)$. Here $s(\varphi)$ is any upper bound for the time to decide the feasibility of any row $\rho$ of length $w=|\varphi|$. Thus the $O((R+1)wsat(\varphi))$ claim will be proven if we can show that 
'$SpMod(\varphi)\cap\rho=\emptyset?$' can be decided in time $\le sat(\varphi)$. Consider $\psi:=\varphi\wedge T(\rho)$ where $T(\rho)$ is the term induced by $\rho$.
Then $\psi\le\varphi$ and $SpMod(\varphi)\cap\rho=SpMod(\psi)$. Since ${\cal C}$ is hereditary and $sat$ is monotone the emptiness of $SpMod(\psi)$ can be tested in time $\le sat(\varphi)$. \quad $\square$

{\bf 9.2.2} It is folklore that the models of a DNF can be enumerated efficiently, but what exactly is meant by 'efficiently'? In 2.5.1 we saw that 'polynomial total time' is easy to achieve using Pivotal Decomposition. What about polynomial delay enumeration? Apparently this was first proven
by Yann Strozecki  in his 2010 Thesis; the proof is reproduced in [2]. Stepping up matters from $Mod(\varphi)$ to $Mod(\varphi,k)$ one can show the following.

\begin{tabular}{|l|} \hline \\
{\bf Corollary 3:} 
If $\varphi$ is given as DNF with $t$ terms then Mod$(\varphi, k)$ can be
enumerated \\(one-by-one, but with potential of compression) in $O((R+1)tw^2)$ time where $R = |\mbox{Mod}(\varphi, k)|$.\\
\hline  \end{tabular}

{\it Proof.} The class ${\cal C}$ of all DNF's is hereditary. If we can verify that for each $\varphi\in{\cal C}$ of arity $w$ one can decide in time $sat(\varphi)=O(tw)$ whether
$SpMod (\varphi) := Mod(\varphi,k)$ is empty, then the claim follows from Corollary 2. So let
  $\{T_1, \cdots, T_t\}$ be the set of terms of $\varphi$. Then Mod$(\varphi) = r(T_1) \cup \cdots \cup r(T_t)$ where $r(T_i)$ is as in 1.1. Hence $r \cap \mbox{Mod}(\varphi, k) \neq \emptyset$ iff some set $r \cap r(T_i)$ contains a $k$-model. Now $r \cap r(T_i) = \emptyset$ iff ones$(r) \cap \mbox{zeros}(r(T_i)) \neq \emptyset$ or zeros$(r) \cap \mbox{ones}(r(T_i)) \neq \emptyset$. If $r \cap r(T_i) \neq \emptyset$ then $\rho_i : = r \cap r(T_i)$ can again be written as $012$-row. Evidently $\rho_i$ contains at least\footnote{One can output {\it all} $k$-models contained in $\rho_i$ in compressed format by using the $g$-wildcard.} one $k$-model iff $|\mbox{ones}(\rho_i)| \leq k \leq |\mbox{ones}(\rho_i)| + |\mbox{twos}(\rho_i)|$. Therefore $sat(w) = O(tw)$. \quad $\square$

  The closest match in the literature to Corollary 3 seems to be the lengthy article [14] which shows that when $\varphi$ is in d-DNNF format\footnote{This deterministic Decomposable Negation Normal Form is due to Adnan Darwiche.}, its $k$-models can be enumerated one-by-one with constant delay. However, neither of d-DNNF and DNF subsumes the other. While d-DNNFs {\it do} subsume BDDs, let us recall from 2.3 that using $g$-wildcards the $k$-models of BDDs can be enumerated in a {\it compressed} format (and in polynomial total time). 
  
  It e.g. follows at once from Corollary 3  that the $k$-faces of a simplicial complex given by its facets can be enumerated in polynomial total time. See [arXiv:1812.02570] for a direct proof, and for an alternative method which cannot boast 'polynomial total time' but offers higher compression in practice. 

\section{Clausal Imposition as a row-splitting mechanism}

In order to theoretically assess the clausal $n$-algorithm informally introduced in Section 4, fix any CNF $\varphi$ consisting of $h=h(\varphi)$ clauses $C_1$ to $C_h$. Let ${\cal R}$ be the set of all 012n-rows $r$ of length $w=|\varphi|$. Define $deg(r)$ as the maximum number $k$ such that each bitstring $u\in r$ satisfies all clauses $C_1,\ldots,C_k$. Hence each row of degree $h$ is final, and so $({\cal R},deg,h)$ is a
row-structure of $Mod(\varphi)$.

By definition the coupled row-splitting mechanism $({\cal A}_1,d,{\cal A}_2,s)$ consists of algorithms ${\cal A}_1$ and 
 ${\cal A}_2$ that calculate degrees and split rows respectively. Upon calculating the costs $d(\varphi)$ and $s(\varphi)$ to do so, the Master Theorem will tell us the overall cost of processing $\varphi$ with the clausal $n$-algorithm.

\begin{tabular}{|l|} \hline \\
{\bf Corollary 4:} Let ${\cal C}$ be a hereditary class of Boolean functions and let $sat:\ {\cal R}\to\N$  \\
 be monotone. Suppose that for each $\psi\in{\cal C}$ it can be tested in time $\le sat(\psi)$ whether\\
  or not $Mod(\psi)=\emptyset$. Then for each $\varphi \in {\cal C}$ with $h$ clauses and arity $w$ one  \\
  can enumerate $Mod(\varphi)$ with $R$ many disjoint $012n$-rows in time $O((R+1)hw(h+sat(\varphi)))$.  \\ \\ \hline 
 \end{tabular}

% Let ${\cal C}$ be a hereditary class of Boolean CNF's  such that the satisfiability of 
%each arity$w$ member of ${\cal C}$ can be tested in time $O(sat(w))$, where $sat$ is a monotone function. 
%Then for each $\varphi \in {\cal C}$ having arity $ w$ and $h$ clauses one can 
%Mod$(\varphi)$ with $R$ many disjoint $012n$-rows in time $O(Rhw(h+sat(w)))$.\\ \\ \hline \end{tabular}

{\it Proof:} By the Master Theorem $Mod(\varphi)$ can be enumerated in time $O(Rh(d+s))$. We will show that $d=d(\varphi)=O(hw)$ and
 $s=s(\varphi)=O(wsat(\varphi))$. This will do the job in view of $O(Rh(d+s))=O(Rh(hw+wsat(\varphi))$.

As to $d(\varphi)$, the degree of a $012n$-row $r$ (i.e. its pending clause) is calculated by scanning the $h$ clauses $C_i$ until $C_i \cap \mbox{ones}(r) = \emptyset$. Hence $d(\varphi) = O(hw)$. 

As to the cost $s(\varphi)$ of splitting feasible, non-final 012n-rows $r$ of length $w$,
imposing a clause $C$ of length $\tau \leq w$ upon  $r$  entails raising an Abraham 0-Flag and Abraham 1-Flag, according to the negative and positive literals of $C$ respectively. This
costs $O(w\tau)= O(w^2)$. Each of the $\tau$ many candidate sons $\rho$ needs  to be tested for feasibility. Testing the feasibility of $\rho$ amounts to testing the satisfiability of $\psi = \varphi \wedge T(\rho)$. Since ${\cal C}$ is hereditary, $\psi$ belongs to ${\cal C}$ and 
 its satisfiability is testable in time $s\leq sat(\varphi)$ by the monotonicity of $sat$. Hence 
 $s(\varphi) = O(w^2 + wsat(\varphi)) = O(wsat(\varphi))$.
  $\square$
	
	As shown in [10, claim (14)], if ${\cal C}$ is the hereditary class of all Horn CNF's then one can achieve $sat(\varphi)=O(hw)$. Hence the cost in Corollary 4 becomes $O(Rhw(h+hw))=O((R+1)h^2w^2)$, which coincides with the cost in [10,Theorem 2].

\section*{References}
\begin{enumerate}
	
\item[{[1]}]  Valiant L.G. (1979), The complexity of enumeration and reliability problems, SIAM J. Comput. 8, 410-421.

\item[{[2]}]  Capelli F., Strozecki Y. (2020),	Enumerating models of DNF faster:
 breaking the dependency on	the formula size, Discrete Applied Mathematics, accepted.

\item[{[3]}] Abraham J.A. (1979), An improved algorithm for network reliability, IEEE transactions on Reliability R-28, 58-61.

\item[{[4]}]  Crama Y., Hammer P.L. (editors) (2011), Boolean Functions, Cambridge University Press.

\item[{[5]}] Rauzy A.,  Châtelet E.,  Dutuit Y., Bérenguer C. (2003), A practical comparison of methods to assess sum-of-products, Reliability Engineering \& System Safety, Volume 79, 33-42.

\item[{[6]}]  Fazel K.,  Thornton M.A.,  Rice J.E. (2007), ESOP-based Toffoli gate cascade generation, PACRIM.

\item[{[7]}] Knuth D. (2012), The Art of Computer Programming, Volume 4A, Addison-Wesley.

\item[{[8]}] Wild M., Compression with wildcards: All $k$-models of a Binary Decision Diagram,\\ arXiv:1703.08511.

\item[{[9]}] Creignou N.,Hebrard J.J. (1997),	On generating all solutions of generalized
satisfiability , Informatique théorique et applications, Vol. 31,  499-511.

\item[{[10]}] Wild M. (2012), Compactly generating all satisfying truth assignments of a Horn formula, J. Satisf. Boolean Model. Comput. 8, 63-82.

\item[{[11]}] Zivny S. (2012), The complexity of valued constraint satisfaction problems, Springer.

\item[{[12]}]  M. Wild, S. Janson, S. Wagner, D. Laurie, Coupon collecting and transversals of hypergraphs. Discrete Math. Theor. Comput. Sci. 15 (2013), no.2, 259-270.

\item[{[13]}] Toda T., Soh T. (2016), Implementing Efficient All Solutions SAT Solvers, J. Exp. Algorithms 21, 44 pages.

\item[{[14]}]  Amarilli A., Bourhis P.,  Jachiet L., Mengel S. (2017), A Circuit-Based Approach to Efficient Eumeration, ICALP.

%\item[{[W5]}] M. Wild, Counting or producing all fixed cardinality transversals. Algorithmica 69 (2014), no.1, 117-129.
%\item[{[W6]}] M. Wild, S. Janson, S. Wagner, D. Laurie, Coupon collecting and transversals of hypergraphs. Discrete Math. Theor. Comput. Sci. 15 (2013), no.2, 259-270.
%\item[{[W7]}] M. Wild, The joy of implications, aka pure Horn formulas: Mainly a survey. Theoretical Computer Science 658 (2017) 264-292.
%\item[{[W8]}] M. Wild, Compressed representation of Learning Spaces, to appear in J. Math. Psychology.
%\item[{[W9]}] M. Wild, Revisiting the enumeration of all models of a Boolean 2-CNF, in the arXiv.
%\item[{[YIH]}] Y. Yamamoto, K. Iwanuma, N. Hidetomo, Practically fast non-monotone dualization based on monotone dualization, arXiv1311.4639.
%\item[{[W1]}] M. Wild, Output-polynomial enumeration of all fixed-cardinality ideals of a poset, respectively all fixed-cardinality subtrees of a tree,  Order 31 (2014) 121-135.
\end{enumerate}

\end{document}